# Improving Energy Efficiency of Oil Platforms Through Optimal Loading of Diesel Generators Using Machine Learning and Search Algorithms.

Khivishta Boodhoo[a], Isaac Triguero[b c d], Josh Plumbly[e], Bruce Nicolson[e], William Meredith[a], Nicholas Watson[f*]

[a] Low Carbon Energy and Resources Technologies Research Group, Faculty of Engineering, University of Nottingham.
[b]Computational Optimisation and Learning Lab, School of Computer Science, University of Nottingham, Nottingham, United Kingdom

[c]Department of Computer science and Artificial Intelligence,University of Granada, Spain

[d]DaSCI Andalusian Institute in Data Science and Computational Intelligence, Granada, Spain

[e]Intelligent Plant, Aberdeen, Scotland

[f]School of Food Science and Nutrition, University of Leeds, Leeds, United Kingdom.

[*]Corresponding Author

## Abstract

With the rising global demand for energy, fossil fuel resources (oil and gas) are becoming depleted and their prices are increasing. Coupled with climate change challenges, it is necessary to use and produce energy efficiently. When it comes to energy production from oil and gas platforms, challenges such as inefficient energy consumption, system failures, accessibility, and their environmental impact, among others, need to be addressed. Amid the Fourth Industrial Revolution, artificial intelligence has emerged as a promising tool to optimise industrialised systems and processes. Notably, machine learning, a subset of artificial intelligence, is being applied to energy systems such as oil platforms because of its capabilities to enhance safety, sustainability and efficiency through the collection and modelling of data. However, the majority of previous studies have focused on increasing oil production, whereas energy consumption on the oil platform itself has not received as much attention. Therefore, this study investigated the use of machine learning and search algorithms to consume diesel more efficiently on oil platforms. To achieve this, data collected over 18 months from an offshore oil platform in Scotland were used, with the primary diesel-consuming equipment being four diesel generators. First, preprocessing of the data through an exploratory data analysis and outlier detection was performed. Machine learning methods for regression problems were employed to predict the daily diesel consumption for different power loads on diesel generators. Multiple Linear Regression and Artificial Neural Networks achieved the best predicted performance over Extra Trees Regressors, Extreme Gradient Boost, and Random Forest. Following the diesel consumption predictions, search algorithms determined possible combinations of daily power loads for a daily optimal total loading on the diesel generators, resulting in minimal diesel consumption. The results provided an average diesel saving of 27 % per day (compared to the worst daily power loads combinations), which is approximately 24000 litres/day demonstrating significant opportunities to save energy on offshore oil platforms.

## Abbreviations[1]

[1] AI: Artificial Intelligence, ANN: Artificial Neural Network, Avg.:Average, DGs: Diesel Generators, DTs: Decision Trees, RF: Random Forest, ETRs: Extra Trees Regressor, MLR: Multiple Linear Regression, ML: Machine Learning, FPSO: Floating Production Storage and Offloading, PLs: Power Loads, PCP: Parallel Plot Coordinate, RMSE: Root Mean Square Error, MSE: Mean Square Error, MAE: Mean Absolute Error, $R_2$: Coefficient of determination

# 1. Introduction and Background

The demand for energy in the modern world is constantly increasing (EIA, 2023), which has led to the development of various means to produce energy. In 2017, fossil fuels produced 64.5% of the electricity worldwide (World Nuclear Association, 2023), and in 2020, oil and gas accounted for 57.3% of the total energy production (electricity, transport, and heat) (Hannah & Max, 2020). Oil and gas are fuel materials, and like coal, they are burned to produce heat or power. Although it is polluting, obtaining energy from crude oil is still integrated into society. The U.S. Energy Information Administration (EIA) estimates that the world consumed 92.2 million barrels per day (b/d) of petroleum and other liquid fuels in 2020 (U.S EIA, 2021). In 2022, approximately 7.4 billion barrels of total U.S. petroleum consumption consisted of 43% motor gasoline (including fuel ethanol), 20% distillate fuel (heating oil and diesel fuel), and 8% jet fuel. Petroleum products are additionally, used in vehicles, heating buildings, and electricity, and as raw materials for plastics, solvents, and multiple other goods (U.S.Energy Information Administration, 2023).

Because of this widespread use and associated costs, efficient fuel usage has been a concern and looking for ways to maximise the effectiveness of fuel consumption while keeping waste and pollution at their lowest is a hot topic of interest. Naturally, optimisation of fuel use is directly linked to the energy efficiency of a system since minimising fuel use while maximising energy output leads to higher system efficiency. This is essential for reducing energy demand, depleting fossil fuels, and damage to the environment. Fuel optimisation has thus been applied in various industries such as transportation, industrial processes, energy generation and buildings while yielding positive outcomes. For instance, Meng and Gustav (2013) researched the adoption of optimal driving strategies using intelligent cruise control to minimise fuel consumption in heavy vehicles. Others have developed route optimisation software claiming to reduce fuel costs by up to 30 % (GSMtasks, n.d.). One study by Metlek, (2023) investigated fuel consumption by predicting engine fuel consumption in aircraft to optimise its usage.

Other studies have explored the prediction of fuel consumption in various transportation contexts. For example, a study conducted by Y.Chen et al., (2017) utilised a Data Driven model (DDM) to predict fuel consumption based on specific vehicle inputs, yielding less than 10% error, but this was limited to only one metropolitan area. Wickramanayake and Bandara, (2016) used such DDMs for long-distance buses, with a particular DDM providing more accurate results after suitable feature selection, though its applicability to other scenarios remains uncertain. Moreover, other studies, such as Chaudhary (2019), Gong et al. (2021), and Hussain et al. (2022) predicted fuel consumption in vehicle systems using diverse inputs, resulting in good accuracy of the prediction models. Another study by H.Zheng et al., (2021) proposed a DDM model for marine diesel engines, achieving high prediction accuracy. Overall, these studies emphasise the need for preprocessing and demonstrate the usefulness of a data driven approach but, at the same time, lack clarity on model usage after development and the broader applicability of such models; other than the specific area that they have been applied to. Nevertheless, past research establishes useful insights for potentially extending the developed principles to diesel consumption on oil platforms.

As such, oil platforms consume a large amount of fuel due to oil extraction, and oil refining leading to the requirement of a substantial amount of energy to operate such platforms. In this context, Diesel Generators (DGs) have long been used to provide electricity on oil platforms, but their efficient operation has received little consideration. Instead, increasing the production volume of oil and gas while minimising the cost of production, a concept known as product optimisation has been investigated in many studies and has remained a priority ever since, in order to increase the value of oil companies and maximise their revenue (Krishnamoorthy et al., 2019). This goal has been fundamental to the industry from the beginning and is deeply rooted in the principles of petroleum engineering (B. Guo et al., 2007) leading to the overlooking of other approaches (e.g., efficient diesel consumption). Employed methods can include investigations to maximise both the total profit and usage availability of a typical combination of gas turbine engines used for power generation in oil and gas production (Houssein, 2010), the efficient and successful utilisation of materials for oil platforms (Bijan et al., 2005) (Y. Guo et al., 2024), optimal production plans and mode of operations (Silva & Camponogara, 2014) of oil platforms. Additionally, traditional model-based strategies for improving the development and production of petroleum fields have largely focused on designing oil platforms and simulations of reservoir engineering (Khor et al., 2017).

The literature also includes numerous studies on oil platforms focused on improving the efficiency of the oil drilling process and oil production to minimise energy use (Hegde & Gray, 2017; Barbosa et al., 2019; Sircar et al., 2021). For instance, Nguyen et al., (2019) conducted a study on developing a new methodology for using oil platform layouts to work on process simulation, energy analysis, and optimisation routines. Svendsen., (2022) evaluated a new integrated energy system model by assessing the effects of modifications in system configurations and operating parameters on oil platforms using the 'Oogeso' open-source optimisation and simulation tool for analysis of energy system operation. Until now, scholars have mainly focused on offshore platform power system networks and power grid stability control technology (Wu et al., 2024), (Camargo et al., 2024), operation parameters, working conditions, the introduction of new sources of energy, and oil and gas treatment processes, control, and energy recovery (Nguyen et al., 2016) (Lou et al., 2019) (Zhang et al., 2019) (Allahyarzadeh et al., 2018)(.Zheng et al., 2022) , (Ribeiro et al., 2024). More recently, Meng et al. (2025) proposed a low-carbon offshore-oilfield power-system planning method that jointly considered wind generation and carbon capture, utilisation and storage, further illustrating the field's emphasis on system configuration and emissions reduction.

However, emerging techniques that include DDMs, such as Machine Learning (ML) models, have recently gained popularity, as highlighted by Choubey and Karmakar (2021). ML techniques commonly employed in the oil and gas industry include Linear Regression (LR), Decision Trees (DTs), Principal Component Analysis, Support Vector Machines and Artificial Neural Networks (ANNs) (Delavar & Ramezanzadeh, 2024) (X. Chen et al., 2024) (Elahifar & Hosseini, 2024). For instance, a study conducted by Al-Mudhafer & Shahed, (2011) used a Genetic Algorithm, a type of ML algorithm, to maximise oil recovery in Iraq by determining the future reservoir performance regarding infill drilling. Another study (Cao et al., 2016) used an ANN to predict the well production performance of existing and nearby wells. A study investigated Carbon Dioxide ($CO_2$) flooding to improve oil recovery by using a Random Forest (RF) algorithm (another type of ML algorithm) and to predict $CO_2$-WAG (Water Alternating Gas) performance with high accuracy and high computational efficiency under conditions of various injection parameters (H.Li et al., 2022).

Although, ML has been seen to optimise the process of producing oil and gas for maximisation oil and gas production, reducing the energy use by offshore oil platforms is also being perceived as of utmost importance (Anthony Arinze et al., 2024) (Saghir et al., 2024) . This is because an offshore oil platform is an isolated energy system, which requires a large amount of electricity, mechanical power, and thermal energy. Each facility on an oil platform (which can be considered as an infrastructure or an installation) can use a few to several hundred MW of energy per day, depending on the petroleum properties, export specifications, and field lifetime (Nguyen et al., 2019). DGs usually supply the energy needed for offshore drilling rigs, and typically use 20–30 $m^3$ of diesel fuel per day, depending on the operations performed (Ipieca, 2013). Estimates suggest that it takes 0.3 to 0.5 gallons of diesel fuel to produce one barrel of oil on an offshore oil platform (U.S. Energy Information Administration, 2023a). Thus, diesel consumption is one of the major contributors to energy use on these platforms (Generator Source, 2023). Monitoring and optimisation of diesel consumption on oil platforms can play a vital role in decreasing energy usage in offshore fields (Worldwide Power Products, 2023).

In fact, some modelling methods have been previously used to study efficiency of DGs on oil platforms. For instance, Yadav et al. (2011) proposed an Improved Harmony Search algorithm for optimal power generation scheduling of eight DGs units. The fuel consumption rate of the oil rig using various optimisation methods, such as the MinMax method, MinCon method, Genetic Algorithm, Particle Swarm Optimisation, Harmony Search and Improved Harmony Search were investigated; with the Improved Harmony Search resulting in an output which had the lowest diesel consumption. Another study by Rao et al. (2015) investigated the variations in the loading of each operating DG and the corresponding specific fuel consumption values, which were plotted against the total load demand for an offshore support vessel. A Genetic-Algorithm-based approach was employed to minimise fuel consumption. The results showed that a vessel with a proper selection of DG capacities can significantly save fuel with a reduction of up to 10.1 % while providing higher operating flexibility and endurance of the DGs owing to operation at optimal loading under most loading conditions. Using real field data from a diesel-electric platform supply vessel, Peixoto et al. (2024) likewise found that optimised generator dispatch and battery-assisted operation could reduce fuel use, although the analysis concerned vessel operating modes rather than an oil platform's daily DG load distribution. More directly, Cemeljic et al. (2024) combined an adaptive power-management algorithm, a mathematical rig model and ML-based event prediction for parallel DG operation on a jack-up drilling rig, reporting fuel savings of up to 1,700 litres per month in economy mode.

However, looking at the optimal methods of these past studies, mainly Improved Harmony Search and Genetic Algorithm, the models may lack robustness as the Improved Harmony Search is sensitive to the initial population, and the Genetic Algorithm may have premature convergence (Khan et al., 2012). Nevertheless, methods that directly optimise diesel consumption through the daily load distribution of DGs on oil platforms remain comparatively underexplored. Yet, various ML algorithms have been used, as previously mentioned, to predict diesel consumption in vehicles and ships, but they have not been applied to oil platforms. As a matter of fact, when considering oil platforms, the focus has almost always been on increasing oil production rather than the energy efficiency of oil platforms because of the rooted perception that more oil means more revenue; besides $CO_2$ emissions were not always a priority, especially during the early days of extracting petroleum. Therefore, insufficient research exists on minimising diesel consumption on oil platforms in past studies nor are end users given enough consideration and information on how to use the developed models. In addition, while a diverse range of ML algorithms has been extensively applied across various research papers in this field of study, there are not enough studies employing both the simplest and most complex ML algorithms to assess and compare their performance in solving a single problem within a research paper.

To fill this gap, the following study explores a novel methodology based on ML and search algorithms for reducing diesel consumption on an oil platform along with a visualisation tool for the end user to identify the optimal load distribution for minimal diesel consumption. To the best of the authors' knowledge, this particular methodology has not been applied to oil platforms in the existing literature.

The objectives of this study were as follows:

i. Analyse the current feature patterns of daily diesel consumption on an oil platform to identify any possible relationships and find the daily usage of diesel and daily power loads on DGs.
ii. Investigate the use of different ML models (from simple to more complex) to predict diesel consumption based on different combinations of DG power loads.
iii. Predict diesel consumption for various distributions of power loads on the four DGs using a search algorithm and investigate the effects of DG loading and its impact on diesel consumption on oil platforms. These findings are represented in the form of an end-user graph consisting of possible combinations of daily energy usage of power loads and the corresponding predicted daily diesel consumption.
iv. Determine the most efficient combination of DG power loads that results in the lowest consumption of diesel.

The remainder of this paper is organised as follows: Section 2 presents the methodology, where Section 2.1 describes the oil platform with an overview of the dataset. Section 2.2 gives a general description of the adopted modelling method. In Subsection 2.2.1, the preprocessing steps required to preprocess the data are described, along with feature engineering. Subsection 2.2.2 describes the model-building process and deployment for the ML models and search algorithm. Section 3 presents the results of the model, the discussion section interprets the results and their implications, and section 4 presents the conclusion that summarizes the findings of the study and suggests directions for future research.

## 2. Methodology

This section begins with an overview of the oil platform studied in this work and a process flow diagram of the flow of diesel to the different equipment. An outline of the relevant datasets is then provided before outlining the different modelling strategies used.

### 2.1. Overview of Oil Platform and Dataset

Floating Production Storage and Offloading (FPSO) oil platforms have gradually become mainstream facilities for the production, processing, storage, and transportation of offshore oil fields (Stelios & Edmundo, 2023). This study investigated an FPSO oil platform where diesel was mainly consumed by the five Wartsila DGs while other equipment such as bow thrusters and fire water pumps used negligible amounts of diesel. Annual operation of the duty fire pumps accounted for about 1.4% while its maintenance tests consumed 1% of diesel, and the maintenance tests of the bow thrusters used 0.6% of diesel. The process flow diagram in Figure 1 shows the different equipment using diesel on the FPSO oil platform. Blended Process Gas (BPG) was separated from crude oil and burned in Process Fired Heaters (PFHs), which have been running on diesel since 2017 due to gas deficiency, with a 24-hr flow rate of 50,520 kg/day. The Inert Gas Generators (IGGs) also burned diesel to create a low-oxygen blanket of gas during cargo offloading, with a flow rate of 10,392 kg/day.

Features such as the oxygen ($O_2$) concentration near the IGG and the exhaust gas temperatures of the PFHs were used to determine if the equipment were on or off, for the calculation of daily diesel consumption. Electricity on the FPSO oil platform was provided by five DGs, however one was broken (DG-C) and thus ignored in this study. The power loads on the DGs were recorded at irregular intervals and were available since 2010 but the overall daily diesel usage on the platform was manually recorded as from 2021. A summary of the dataset used in the modelling process is presented in Table 1.

Table 1. Overview of the dataset for the oil platform showing the features selected for this study.

| | **Recorded diesel consumption (Label)** | **Power Loads for Wartsila Diesel Generators (DGs) (Features)** |
|---|---|---|
| **Features** | Recorded diesel usage | Power loads for DGs: A, B, D, E |
| **Data period** | ~ 06-2021 to 12-2022 | ~ 01-2010 to 12-2022 |
| **Timestamps when data is recorded** | The time the value is recorded is at either 23:00 or 24:00 daily | When in use, values recorded at intervals varying between 1-5 mins. When not being used, values are recorded as 0 or at hourly intervals. |
| **Units** | m3/day | kW |

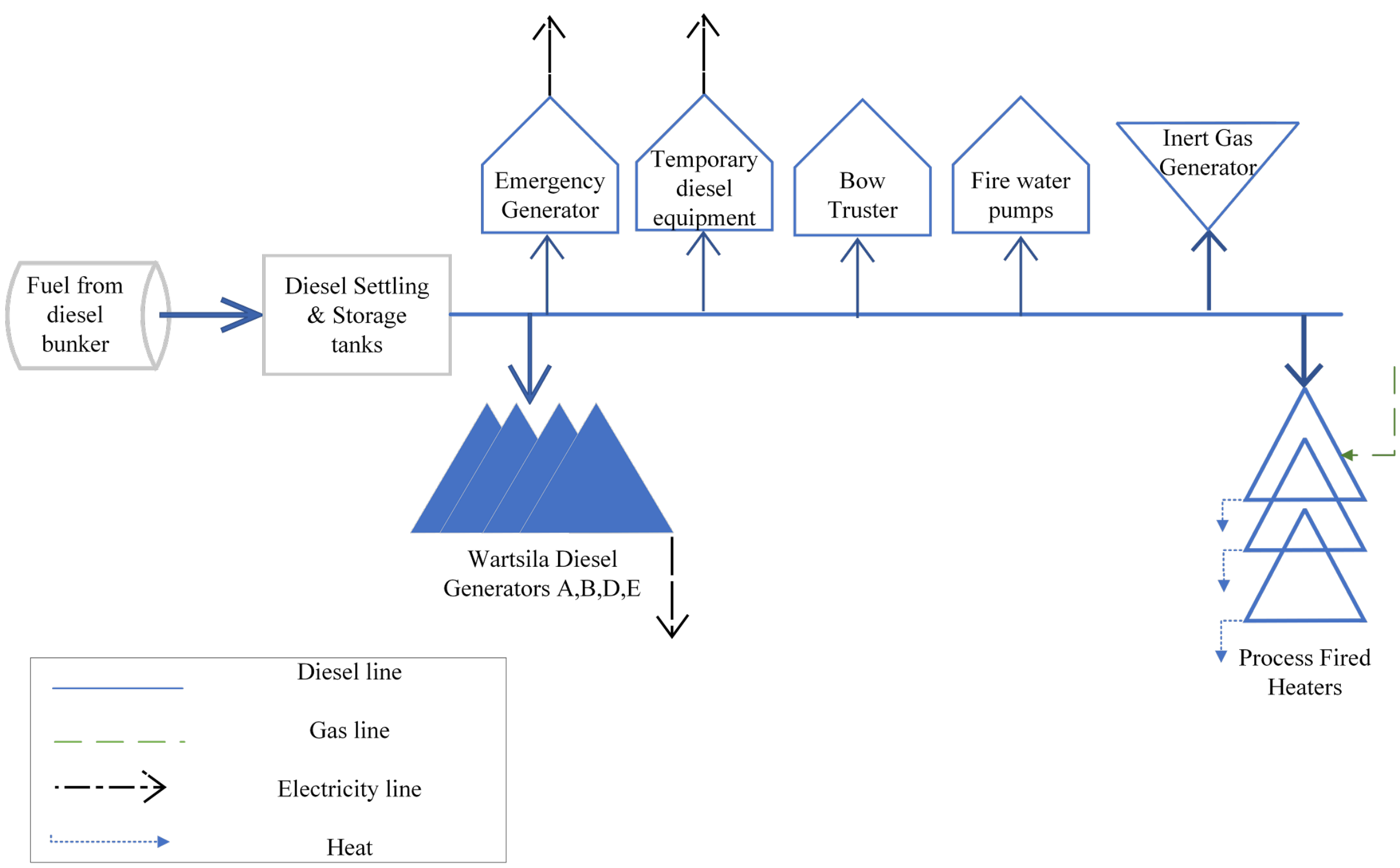

Fig. 1. Process Flow Diagram of FPSO Oil Platform showing the pipeline to different equipment consuming diesel on the selected oil platform.

## 2.2. Method Designed to Reduce Diesel Consumption on the Oil Platform

This section provides an overview of the procedures used in the study. Figure 2 highlights the different steps involved in the methodology for predicting diesel consumption. The calculation of key features, such as daily power loads of the DGs and daily consumption of diesel for PFHs and IGGs, are the first steps in the methodology developed.

As seen in Figure 2, preprocessing steps such as visualisation and outlier removal were first applied. Subsequently, modelling started using a simple statistical model, namely Multiple Linear Regression (MLR). Models of intermediate complexity, such as Extreme Gradient Boosting (XGBoost), Random Forest (RF), and Extra Trees Regressors (ETRs), have been developed. Finally, a more complex ML model, an ANN, was employed. After training, testing, and validating the ML models for accurate diesel consumption predictions, the model performance metrics determined the best models. Generated inputs of daily combinations of power loads on the DGs, were produced using a search algorithm and used in the best-performing ML models. The resulting diesel predictions from the corresponding generated power load combinations were then visualised in a multidimensional plot to identify the best daily combinations of power loads producing the least diesel consumption.

Data

Feature Engineering.
Exploratory Data Analysis (EDA).
Stacked bar chart and scatter plots for data analysis and outlier detection.
Correlations analysis before preprocessing.
Mahalanobis Distance (MD) for outlier removal.

Modelling

Regression and ML models to predict diesel consumption:
Multipe Linear regression (MLR).
XGBoost.
Extra Trees Regressors (ETRs).
Artificial Neural Networks (ANNs).

Deployment

Generation of selected combinations of daily loads for model deployment.
Predictions of diesel consumption for the generated daily loads.
Finding the least, most and median diesel consumption and their corresponding daily load combinations.
Result visualisation through Parallel Coordinate Plot (PCP).

Fig. 2. Flowchart representing the different preprocessing steps in this study.

### 2.2.1. Data Preprocessing

This section covers the data preprocessing steps used in this study. These include feature engineering, Exploratory Data Analysis (EDA), outlier detection and removal, data visualisation before and after outlier detection and removal, correlation analysis, and analysis of feature patterns and relationships.

#### 2.2.1.1. Feature Engineering

The features of the dataset must be aligned in granularity with the model's output label, which in this case was daily recorded diesel consumption. Power loads on the DGs, captured at varying intervals (secs or mins), were converted into average daily load values by establishing the relationship between power loads and daily average energy used/required based on

$$E_{(kWh)} = P_{(kW)} \times T_{(h)} \quad (1)$$

kWh = kilowatts × hours

where E is the average energy used/required per day, P is the power load of the DGs, and T is the operating time. From further calculations, the PFHs were found to consume a constant amount of diesel between 29-12-2017 and 29-12-2022 which is approximately 59 $m^3$/day. Since this value remained constant every day, this feature had no useful relationship with the target feature, as when comparing this diesel consumption with the daily total diesel consumption resulted in a weak correlation. Hence, diesel consumption by PFHs was a redundant feature (Pudjihartono et al., 2022) and could be ignored in the succeeding calculations. From 29-12-2017 to 29-12-2022, the amount of diesel consumed by the IGGs was an average of 1.3 $m^3$/day, which was approximately 1.8 % of daily total diesel consumption. Subsequent comparison of the values of daily diesel consumption of the IGGs with the daily total recorded diesel consumption also demonstrated a weak correlation; therefore, these values could also be ignored in further modelling. Furthermore, because the label of the ML model was daily diesel consumption and these values only started from 01-06-2021, the data range acquired after data preprocessing for all features and label was from 29-06-2021 to 29-12-2022 (500 rows). A seasonality feature was added to the dataset by hot encoding the different months into numbers representing each season, for example, summer: 1, autumn: 2, spring: 3, and winter: 4. Moreover, the best-performing ML model was trained, tested, and validated with and without the seasonality feature to determine whether there were any differences in the model performance metrics when predicting the daily diesel consumption on the test set.

#### 2.2.1.2. Exploratory Data Analysis

Exploring the dataset through data visualisation helped to identify any correlations between features and determine any data preparation required for effective modelling. A stacked bar chart was plotted to visualise the relationships between the daily average energy used/ required by DGs A, B, D, and E (also known as the daily average loads) and the daily diesel consumption (Figure 3). Some recorded daily diesel consumption data points varied linearly with the daily total average stacked loads (daily total energy used/required), whereas other data points did not appear to correlate with the daily total average stacked loads (daily total energy used/required), with some even exhibiting rather erratic behaviours. These data points required further investigations to understand if they were outliers, as shown in Figure 3.

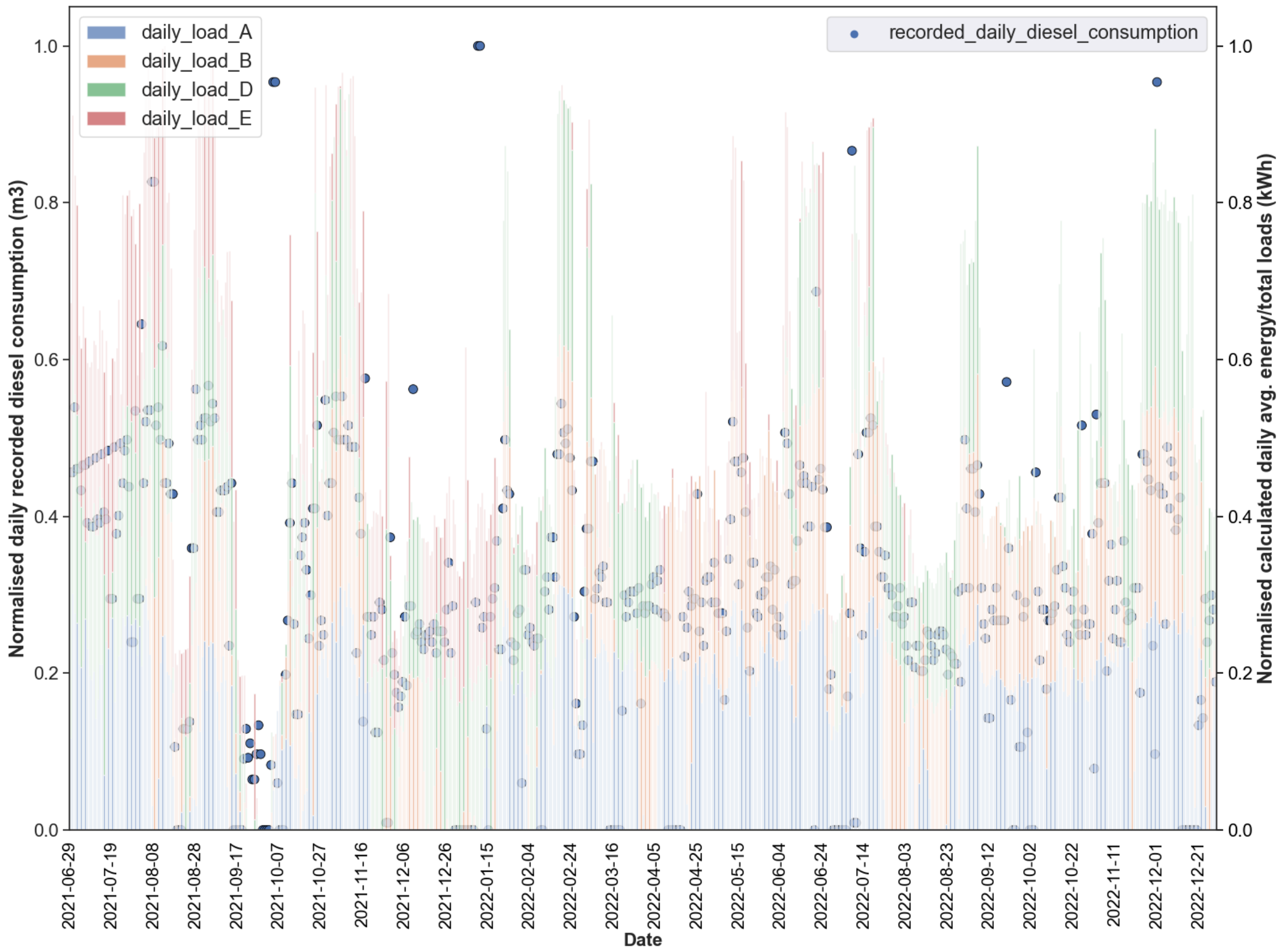

Fig. 3. Normalised daily recorded diesel consumption(m3) as the against date with deliberate pale coloured bars to capture the variations in daily recorded diesel consumption with variations in the power loads.

Moreover, correlations between the daily average total loads (daily energy used/required) and the daily recorded diesel consumption were investigated using the Spearman correlation statistical method, which evaluates the direction and strength of the association between the two variables (The Concise Encyclopedia of Statistics, 2008). It is a non-parametric measure; hence, it makes no assumptions about the data distribution but it assesses the rank order of the data. Spearman correlation was used to determine the relationship between the daily total loads and the daily recorded diesel, resulting in a value of approximately 0.68. As the correlation was not high, other preprocessing steps were required, as outlined in the next Section 2.2.1.3.

#### 2.2.1.3. Outlier Detection and Removal

Abnormal data points were removed when there was a significant increase in the daily average total loads (daily energy used/required), yet the daily recorded diesel consumption remained approximately zero, which means that there were some potential sensor malfunctions or data recording errors. Similarly, when the daily average total loads (daily energy used/required) remained zero, but the daily recorded diesel consumption values increased, the data points were further removed under the same circumstances and assumptions. Next, the daily average total loads and the recorded daily diesel consumed were plotted in Figure 4 where outliers were identified and removed using Mahalanobis Distance (MD) (De Maesschalck et al., 2000). MD considers the covariance structure of the data, making it more sensitive to outliers compared to distance metrics that neglect this information (The Concise Encyclopedia of Statistics, 2008).

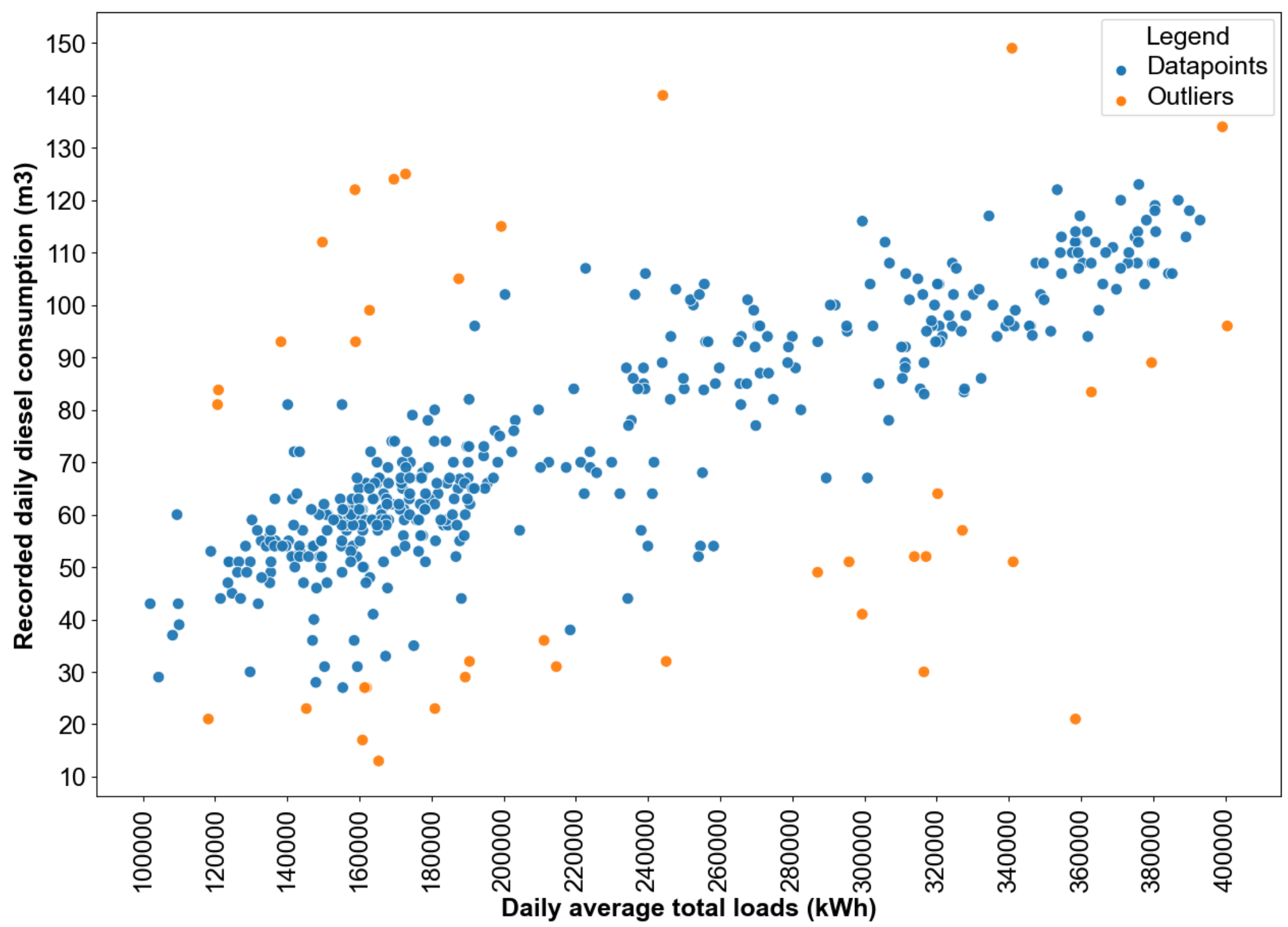

Fig. 4. Recorded daily diesel consumption against daily average loads for data points used for modelling (blue) and outliers removed from the dataset (orange) scatterplot.

### 2.2.1.4. Heatmap for Data Visualisation After Preprocessing

A resulting heatmap was plotted in Figure 5 to analyse the correlations of the features after choosing the features for the model inputs and preprocessing steps to determine if they were effective or not.

The heatmap showed a strong correlation between the daily average total loads and daily recorded diesel consumption with a correlation of 0.87. Daily average total power loads and daily average power load A showed a relatively high correlation of 0.72 which meant that it was the most used among the other DGs' power loads. Daily average power load A also had a correlation of 0.61 with the daily recorded diesel use. PFHs average daily diesel consumption and daily IGG average diesel consumption had a correlation of 0.05 and 0.038 respectively with the daily total recorded diesel consumed confirming that they did not play a large role in the consumption of diesel on the oil platform as previously stated and therefore had not been considered as inputs for the models developed in this work.

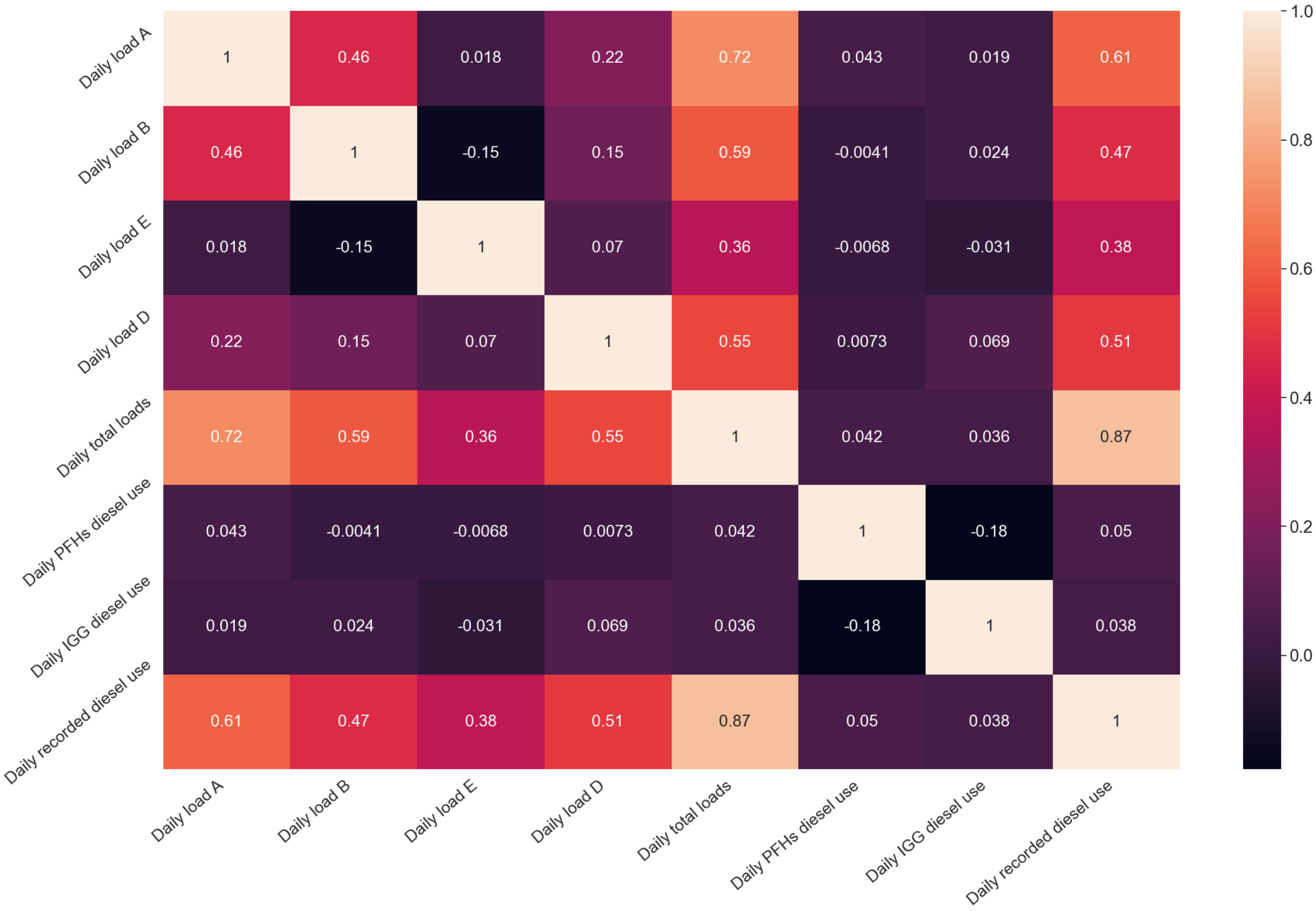

Fig. 5. Correlations between the selected features and target label of the oil platform dataset after preprocessing, with numberings and colours showing the strength of the correlations.

#### 2.2.1.5. Data Normalisation

The inputs to the model were daily average loads A, B, D, and E, and the seasonality feature which was added in the first training, testing and validation of the ML models and subsequently removed in another session of training, testing and validation to see if there were any notable different in the model performance metrics due to seasonal variations. The target label of the model was the recorded daily diesel consumption. All input features were normalised using the min-max function (Patro & Sahu, 2015) to help reduce the scale of the input and output features over a certain range making it easier for the subsequent algorithm to work more efficiently.

#### 2.2.1.6. Analysing Feature Patterns to Identify Useful Relationships.

The input features in previous research, studying diesel consumption in transport, varied greatly from the input features on offshore oil platforms used in this case study from which predictions of diesel consumption were made. Spearman correlation analysis demonstrated the significant impact of daily power loads on daily diesel consumption, with an initial value of 0.67 and 0.87 post-preprocessing, highlighting their linear relationship and the role of data preprocessing.

Looking at the data preprocessing techniques carried out in the predictions of fuel consumption for vehicles and ships as outlined in the literature review (Wickramanayake & Bandara, 2016) (Ahlgren & Thern, 2018) (Chaudhary, 2019) (Gong et al., 2021) (H. Zheng et al., 2021), resampling, feature selection, removal of abnormalities and outliers and data standardisation techniques, for example through RF, SAS Enterprise miner, quartile method, were some of the methods used. Compared to those studies, this work also employed preprocessing techniques following similar steps such as resampling, feature selection, removal of outliers and data normalisation, although the specific methods may have differed. The removal of abnormalities, for example, was performed using a data visualisation technique, outliers were removed using the MD method, feature selection was conducted through diesel consumption calculations for different equipment and data standardisation was performed through the min and max scaler methods. This showed that the preprocessing methods may vary but similar essential steps are required to prepare the data for modelling. Notably, the model performance metrics after running the model have been found to remain consistent with or without the seasonality feature, indicating minimal weather influence on daily diesel consumption for short durations.

### 2.2.2. Machine Learning Modelling

Five ML models of three different types of ML models were explored investigating their structure, parameters, and training process. The first type selected was a simple statistical model which is an MLR. It starts with a hypothesis between the input and output variables which would be an assumption of a linear relationship between the input features and the target feature (Sinharay, 2010). The selected second type of ML models were tree-based ensemble models namely an XGBoost, RF and an ETRs. These subsequent ML models can automatically discover any underlying patterns without necessarily starting with an assumption. Moreover, they were chosen because they are robust and can still make accurate predictions even when there is some degree of randomness or error in the data (Mienye & Sun, 2022). They are also suitable as the dataset is not relatively a big one. Compared to the three tree-based ensemble models, the third type of ML utilised was a more complex model, which was an ANN model. It was selected as it would be more flexible to learn any kind of intricate relationships which could be achieved by adjusting the weights assigned to the connections between its neurons. During the training phase of an ANN, these weights are adjusted iteratively to minimise the difference between the ANN predictions and the actual outcomes, fine-tuning the ANN's ability to make accurate predictions (Walczak & Cerpa, 2003).

Overall, all the models chosen are well suited for a relatively small dataset and are mostly resilient and appropriate for regression problems (T. Chen & Guestrin, 2016) (Hameed et al., 2021). If the relationships between the DGs and diesel consumption were intricate, complex models would be the best suited to model these relationships since complex models can identify underlying patterns when the data has a high degree of noise or variability whereas if the relationship turned out to be linear and simple, the simplest models chosen could be referred to (Tufail et al., 2023). To train the models, the dataset was first split into training and testing subsets of 0.7 and 0.3 respectively for all the models as shown in Table 2, which is commonly practiced in ML literature (Eliane Birba, 2020). The random state was set to a specific seed value to ensure consistent and reproducible random splits. Additionally, validation was performed to minimise overfitting and maximise the model's ability to perform accurately on future data (Dangeti, 2017). One popular approach to validation is cross-validation, where the training dataset is divided into K folds, and subsequently, k iterations of training and validation are implemented such that within each iteration a different fold of the data is held out for validation while the remaining k − 1 folds are used for learning (Refaeilzadeh et al., 2009). Cross Validation (CV) was performed during grid search where for each combination of hyperparameters in the grid, the model was trained on a subset of the training data and validated on a different subset. This process was repeated k times and the average performance was used to assess the model's effectiveness and find the best combinations of hyperparameters.

Table 2. Splitting dataset according to training, testing and validation sets.

| **Dataset** | **Ratio** |
|---|---|
| Train size | 0.7 |
| Test size | 0.3 |
| Random state | 42 |
| K-folds cross-validation | 10 |

#### 2.2.2.1. Multiple Linear Regression model

MLR is a statistical technique used to model the relationship between a dependent variable and multiple independent variables. It assumes a linear relationship between the dependent variable and each of the independent variables. The general form of multiple linear regression can be expressed as (J. D. Jobson, 1991) :

$$y = b_0 + b_1x_1 + b_2x_2 + ... + b_nx_n + e \tag{3}$$

where y is the dependent variable, $x_1$, $x_2$, ..., $x_n$ are the independent variables, $b_0$ is the intercept, $b_1$, $b_2$, ..., bn are the coefficients for the independent variables, e is the error term. Sklearn.linear_model (Sklearn, 2023a) is a module within the scikit-learn library and has the linear regression model which was used in this work. The cross_validate function in scikit-learn library takes the model, the inputs, and a scoring function as inputs, and performs 10-fold cross-validation.

#### 2.2.2.2. Extreme Gradient Boosting

XGBoost is a common open-source ML library (XGBoost Developers, 2022) that uses gradient-boosting algorithms to train decision trees. XGboost is an ensemble learning method that finds the best tree model. The general equation for XGBoost can be written as (T. Chen & Guestrin, 2016):

$$y = \Phi(x) = \Sigma f_j(x) \tag{4}$$

where y is the predicted target variable, X is the input feature vector, $\Phi(x)$ is the function that predicts the target variable based on the input features, $f_j(x)$ is a weak learner, such as a decision tree, that is used to predict the target variable. An XGBoost model with the hyperparameters in Table 3 was constructed after a grid search was performed to find the best hyperparameters.

Table 3 Hyperparameters of Extreme Gradient Boosting (XGBoost)

| Hyperparameters | Values |
|---|---|
| n_jobs | -1 |
| objective | Reg squared error |
| eta | 0.075 |
| Max_depth | 7 |
| Min_child_weight | 25 |
| subsample | 0.9 |
| Colsample_bytree | 0.7 |
| Eval_metric | rmse |
| Explain_level | 2 |
| K_folds cross-validation | 10 |

#### 2.2.2.3. Random Forest

A RF is an ensemble learning approach consisting of multiple decision trees. The output of the RF is the average of each prediction tree. It is more accurate than a single tree, normally better at handling missing data and eliminates the problem of overfitting which can arise when using a decision tree. In a RF, the node selection is carried out by randomly choosing a subset from the current feature set and thereby choosing the best feature in the sub-feature set. The equation is given (Malakouti, 2023) as below:

$$RF = \frac{1}{k}\sum_{k=1}^{k} h_k(x) \tag{5}$$

where k is the number of separate regression trees generated for the bootstrap samples, x is the input vector, $h_k(x)$ is the mean of predictions made by k regression trees. An RF model with the hyperparameters in Table 4 was constructed after a grid search was performed to find the best hyperparameters.

Table 4. Hyperparameters of Random Forest (RF)

| Hyperparameters | Values |
|---|---|
| Min_samples_split | 5 |
| Max_depth | 10 |
| n-estimators | 100 |
| K_folds cross-validation | 10 |

#### 2.2.2.4. Extra Trees Regressors

ETRs (Sklearn, 2023b) build multiple DTs and aggregate their predictions to make a final prediction. The process of building and aggregating decision trees is repeated multiple times to create an ensemble of trees. The final prediction is then made by averaging the predictions of all the trees in the ensemble. The ETRs approach is an advanced technique derived initially from the RF model where the algorithm constructs a collection of unpruned decisions or regression trees. The equation for the ETRs is as follows (Hameed et al., 2021):

$$y = \Phi(x) = (1/n) * \sum f_j(x) \tag{6}$$

where y is the predicted target variable, X is the input feature vector, Φ(x) is the function that predicts the target variable based on the input features, f_j(x) is the j-th decision tree that predicts the input feature vector x, n is the total number of decision trees in the ensemble.
Table 5 shows the optimised hyperparameters for this model.

Table 5. Hyperparameters of Extra Trees Regressors (ETRs).

| Hyperparameters | Values |
| --- | --- |
| Max features | 0.5 |
| Min_samples_split | 10 |
| Min_samples_leaf | 1 |
| Max_depth | 10 |
| n-estimators | 100 |
| K_folds cross-validation | 10 |

#### 2.2.2.5. Artificial Neural Network

An ANN is composed of interconnected artificial neurons organised into layers. The input layer receives input data, and subsequent hidden layers perform calculations on the data to extract features and identify patterns. The output layer produces the final prediction or classification.
The mathematical equation for an artificial neuron is:

$$y = f(w_1x_1 + w_2x_2 + ... + w_n*x_n + b) \tag{7}$$

where y is the output of the neuron, $x_1, x_2, ..., x_n$ are the inputs to the neuron, $w_1, w_2, ..., w_n$ are the weights associated with the inputs, b is the bias term, f is the activation function that transforms the input signal into the output signal. The ANN developed in this work is defined with two hidden 'Dense' layers, a 'dropout' layer, and a final 'Dense' output layer with a 'sigmoid' activation function in TensorFlow API (TensorFlow, 2023). The model was compiled with an Adam optimiser, Mean Squared Error (MAE) loss, and MSE metric. A grid search was conducted with cross-validation of 10-folds using a Keras wrapper (Keras, 2023). Table 5 shows the optimised hyperparameters of the model.

Table 6. Hyperparameters of Artificial Neural Network (ANN).

| **Hyperparameters** | **Values** |
|---|---|
| 1st dense layer neurons | 30 |
| 2nd dense layer neurons | 20 |
| Input layer activation function | Relu |
| Dropout | 0.01 |
| epochs | 150 |
| Batch size | 10 |
| K-folds cross-validation | 10 |

#### 2.2.2.6. Model Performance Metrics

All the ML models were evaluated using the testing set through several model performance metrics. The model performance metrics used are as follows (Chicco et al., 2021):
$R_2$ calculates the amount of variance in the predictions, explained by the dataset. $R^2$ can range from 0 to 1 and the closer it is to 1, the better the goodness of fit.

$$R^2 = 1 - \frac{\sum_{i=1}^{n} (Y_i - \hat{Y}_i)^2}{\sum_{i=1}^{n} (Y -{}_i Y_i)^2} \tag{8}$$

Mean Absolute Error (MAE), along with its normalised version (NMAE), is the mean of the magnitude of difference between the predicted and actual values and is domain-specific. Ideally achieving lower values in the range of the actual values to be predicted is most desirable.

$$MAE = \frac{1}{n} \sum_{i=1}^{n} | Y_i - \hat{Y}_i | \quad (9)2$$

Root Mean Squared Error (RMSE), and its normalised version (NRMSE) is the standard deviation of the prediction errors and ideally should be as low as possible. MSE is the measure of the average squared difference between the actual and predicted values.

$$RMSE = \sqrt{\frac{1}{n}\sum_{i=1}^{n}(Y_i - \hat{Y}_i)^2} \tag{10}$$

$$MSE = \frac{1}{n}\sum_{i=1}^{n}((Y_i - \hat{Y}_i)^2 \tag{11}$$

where $Y_i$: ground-truth value, $\hat{Y}_i$: predicted value from the model, n: number of datums. The lower the MAE, RMSE and MSE the better a model fits a dataset (best value = 0; worst value = $+\infty$).

### 2.2.3. Deployment

In this section, the use of the ML model (see section 2.2.3.5) with the highest $R^2$ and lowest errors, which was trained and validated using real-world datasets, is explored to determine the optimal loadings for the DGs which would achieve minimal diesel consumption. To be able to use the developed ML model, inputs are generated which consist of various combinations of load distributions amongst the four DGs A, B, D, and E. This dataset is generated by first identifying the range of these inputs using a visualisation method of the original dataset used to train and test the model. This range is then divided to generate different percentages of daily loads values on each DG. Different combinations of daily loads are then produced using Cartesian product. The daily total loads for each combination of loads are also calculated. The ML model is then used to predict the daily diesel consumption for the different combinations of daily loads generated. A brute force approach was used to iteratively find the best combination of daily loads on the DGs having the least daily diesel consumption among the ones having the same daily total load values. The best, median and worst combinations of daily loads depending on the diesel consumption are then visualised in a multidimensional plot and diesel consumption savings are also subsequently calculated.

#### 2.2.3.1. Procedures for Model Deployment

The distribution range of the daily loads for each DG in the original dataset is presented in Figure 6. The plot displays a box (the interquartile range, or IQR) that represents the middle 50% of the data and a line inside that box that indicates the median. Typically, any data points outside of the whiskers (lines extending from the box) are displayed as individual points, and the whiskers illustrate the range of the data (outliers). In this case, no outliers can be seen as they have already been removed.

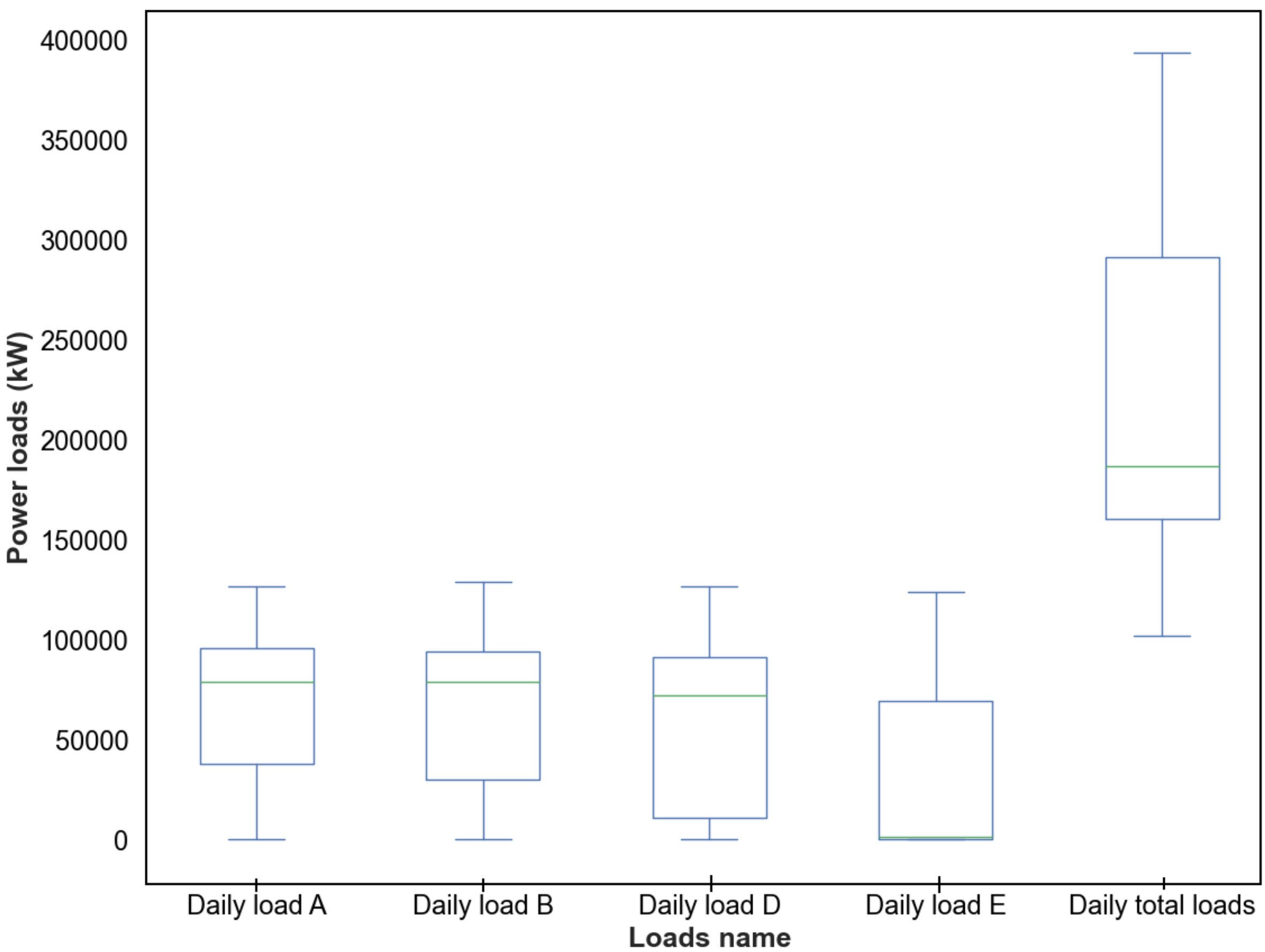

Fig. 6. Box and plot with daily power loads against Diesel Generators (DGs) loads names, showing the maximum, minimum and median of the individual and total daily power loads.

A dataset of potential inputs is then created between the min and max power loads for each DG and this range value is divided into steps to generate percentages of loads as applied on each DG. Thus, different daily load values at an interval of 3% are generated for each DG in between the min and max range of power load values for each DG to represent the possible power load values on each DG. The interval of 3% was chosen as the number of values that can be created is limited due to computational resources. Using the cartesian product (Halpin & Morgan, 2008), all possible combinations of the different power load values produced for each of the DGs A, B, D, and E are calculated. The Cartesian product of input iterables works as a nested for-loops that go through all possibilities. This nested loop structure systematically generates all combinations by iterating through each element in each iterable (Dorst & Van Overveld, 2009). All the possible combinations of daily loads that are likely to exist on a given day on the oil platform are then arranged in a tabular form.

Moreover, the daily total loads were calculated for each of the different combinations produced for daily loads on DGs A, B, D, and E. According to the original dataset, the daily total loads do not have values below 68,000 kW and above 400,000 kW. The generated total daily loads were thus filtered according to the original dataset thresholds of daily total loads. The resulting generated daily load inputs consisted of 1,230,288 different combinations for the daily loads of the DGs.

Each combination of daily loads for DGs A, B, D, and E are used as inputs in the previously developed best performing ML model after being normalised using min-max scaler normalisation. Each combination of daily loads on DGs A, B, D, and E outputs a certain amount of diesel consumed as predicted by the model which is then denormalised to provide actual diesel consumption usage in $m^3$. Multiple different combinations of daily loads on the DGs A, B, D, and E which equate to the same value of daily total loads had a different amount of predicted daily diesel consumed. This is because the DGs vary in efficiency and the optimum amount of diesel used would depend on the correct loading of each of the DGs.

A brute-force approach was then used which is a straightforward and exhaustive algorithmic technique where all possible solutions are considered and checked to find the optimal or desired solution. Brute force searches are simple, reliable, versatile, flexible, and robust which is why they are used in this case to develop the dataset (JavaTpoint, 2021). By exploring all combinations of daily loads on DGs A, B, D, E which equate to the same amount of daily total loads, the daily power loads combinations producing the least amount of diesel for a particular amount of daily total loads can be found through the brute force approach.

The best, worst and median combinations were identified, out of the multiple different daily load combinations that were generated. The best combination of daily loads would highlight the scenario where the DGs are using the least amount of diesel daily. The median is the values of the daily load combinations above and below which half (50%) of the observed data falls, and so represents the midpoint of the multiple combinations available for the same daily total loads. The worst combinations of daily loads could be used to calculate the amount of diesel saved as this represents the most amount of diesel that could be consumed during that particular day, in the worst case scenario.

The selected load combinations are plotted in a Parallel Coordinate Plot (PCP) where each variable is represented by a vertical axis, and each variable's values are connected across all the axes by a line. To help identify patterns or clusters in the data, the lines are coloured. The average daily diesel savings can be calculated from the best and worst combinations of power loads on the DGs for the same total daily loads. The predicted diesel consumption from the best-case combinations was subtracted from the predicted diesel consumption for the worst-case combinations of daily loads on DGs A, B, D, and E resulting in an average daily fuel savings.

## 3. Results and Discussion

This section identifies the best performing ML model for predicting daily diesel consumption based using various power loads combinations. Additionally, it addresses the visualisation for the end user through a multidimensional plot representing power loads combinations which are of interest and evaluates the potential energy savings associated with specific daily power load combinations on the DGs.

### 3.1. Best Performing Machine Learning Algorithms

After training, testing and validating the different ML models using the preprocessed features and label (average daily loads on DGs and daily diesel consumption respectively) from the original dataset, it was found that the MLR and ANN models performed better than the other models (XGboost, RF and ETRs). The MLR model has an $R_2$ score of 0.80 and MAE of 0.08 and the ANN has an $R_2$ score of 0.81 and an MAE of 0.07 as shown in Table 6.

Table 6. Model Performance Metrics for the Multiple Linear Regression, XGBoost, Extra Trees Regressors, Random Forest and Artificial Neural Network models predicting daily diesel consumption.

| **Multiple Linear Regression (MLR)** | |
|---|---|
| MSE | 0.01 |
| RMSE | 0.10 |
| MAE | 0.08 |
| $R_2$ | 0.80 |
| **Extreme Gradient Boosting (XGBoost)** | |
| MSE | 0.013 |
| RMSE | 0.12 |
| MAE | 0.09 |
| $R_2$ | 0.73 |
| **Extra Trees Regressor (ETRs)** | |
| MSE | 0.011 |
| RMSE | 0.10 |
| MAE | 0.08 |
| $R_2$ | 0.78 |

| Random Forest (RF) | |
|---|---|
| MSE | 0.015 |
| RMSE | 0.12 |
| MAE | 0.09 |
| $R_2$ | 0.69 |
| **Artificial Neural Network (ANN)** | |
| MSE | 0.009 |
| RMSE | 0.09 |
| MAE | 0.07 |
| $R_2$ | 0.81 |

MSE = Mean Absolute Error, Root Mean Squared Error = RMSE, MAE = Mean Absolute Error, $R^2$ = Coefficient of determination.

The ANN model had the lowest MSE, RMSE and MAE and the highest $R^2$ score, indicating that it performed better than the other algorithms. The relatively high $R^2$ score for the ANN means that it can explain 81% of the variability in the dataset after rigorous data preprocessing. Moreover, the MLR model estimated the coefficients of the linear equation using least squares regression, while in ANNs, the weights of the connections between the neurons were adjusted during the training process to minimise the MSE. Therefore, the model performance metrics of the ANN indicated that the ANN is better at capturing non-linear relationships than the other algorithms. The RF algorithm had the highest MSE, RMSE and MAE along with the lowest $R^2$ score, indicating the worst performance among the ML methods studied. The XGBoost model had a lower $R^2$ score than the MLR and ETRs models, but it had a similar MSE and RMSE, indicating that it may be better at capturing non-linear relationships between features. These results provide valuable insights into the trade-offs between model complexity and predictive performance. Additionally, it has been noted that the model performance metrics did not change with the seasonal feature when included in the training, testing and validation.

An attempt was made to compare the models' performance developed in this work with other research. Studies such as Wickramanayake & Bandara, (2016) had an MAE of 0.023 for its best performing RF model compared to their ANN and XGBoost for predicted diesel consumption in vehicles. Chaudhary, (2019) reported an $R_2$ of 0.78 for its best performing MLR model under different inputs for the prediction of diesel consumption in vehicles. Gong et al., (2021) achieved a higher prediction accuracy of 81 % for diesel consumed when using their RF model in diesel trucks in comparison to their DTs and ANN. These represent various application areas but exhibit closely related model performance metrics. However, other past works such as (Şahin, 2023) showed an ANN predicting engine performance according to predicted diesel consumption with an $R^2$ of 0.95, Z. S. Chen et al., (2023) found an RF as most accurate with low MAE predicting fuel consumption in shipping and Xie et al., (2023) study which reached an $R^2$ of 0.99 for fuel prediction in the same application area. Consequently, it is evident that the best chosen model varies according to the circumstances. The model performance metrics for different studies exhibit differences and depend on the datasets used, the complexity of the model and the scope of the problem, rendering such comparisons challenging and unreliable.

When considering the best performing models that is ANN and MLR, a choice must be made as to which model to use for deployment. When considering not only the $R^2$ score but also MSE, RMSE and MAE which were lowest for the ANN, the ANN would be the right ML model selection. Considering these factors and their potential impact on future datasets, particularly longer periods and additional inputs (given that another DG was compromised during this study), which may influence the model in a non-linear manner, the decision to use the ANN is further reinforced. ANNs are more flexible than MLR in terms of the types of functions they can learn and the complexity of the relationships they can model between the input and output variables (Wang et al., 2022) (Lendo-Siwicka et al., 2023). ANNs can also process a wider range of functions and capture intricate, non-linear relationships between the input and output variables making them more adaptable (M. Li et al., 2019) than simpler methods such as MLR.

Although ANNs can map more complex relationships it is important to note that an ANN may require more computational power and more time to train than a simpler model. In this case, the MLR took 0.1 sec compared to the ANN which took 7 mins. Depending on the requirements of the user, the models to be deployed can be chosen appropriately. In this case, the ANN was chosen as the model to be used for deployment.

### 3.2. Selected Power Load Combinations and Corresponding Diesel Consumption

The PCP in Figure 8 consists of several axes in descending order which can be described as follows:

1) The predicted amount of diesel used per day.
2) Range of simulated total power loads of DGs namely A, B, D and E per day that have been calculated to be realistically possible.
3) Simulated range of power load per day for DG A.
4) Simulated range of power load per day for DG B.
5) Simulated range of power load per day for DG D.
6) Simulated range of power load per day for DG E.

Figure 8 shows the results for the combinations that used the least, medium and most amount of diesel for a required daily load. The plot in Figure 8 can be used to identify the best, median and worst daily load combinations for the different DGs and a required daily load, which can be displayed separately when further highlighted in Figure 9. The best combination displays the least amount of diesel consumed per day for a required total load, the median line will represent the average amount of diesel used per day, which will be located in the middle of two other lines and the worst combination of the DGs will display the most amount of diesel used per day. Considering only the combinations of DGs A, B, D and E which consumed the least amount of diesel, the results for 88 different daily total loads required are shown in Figure 10.

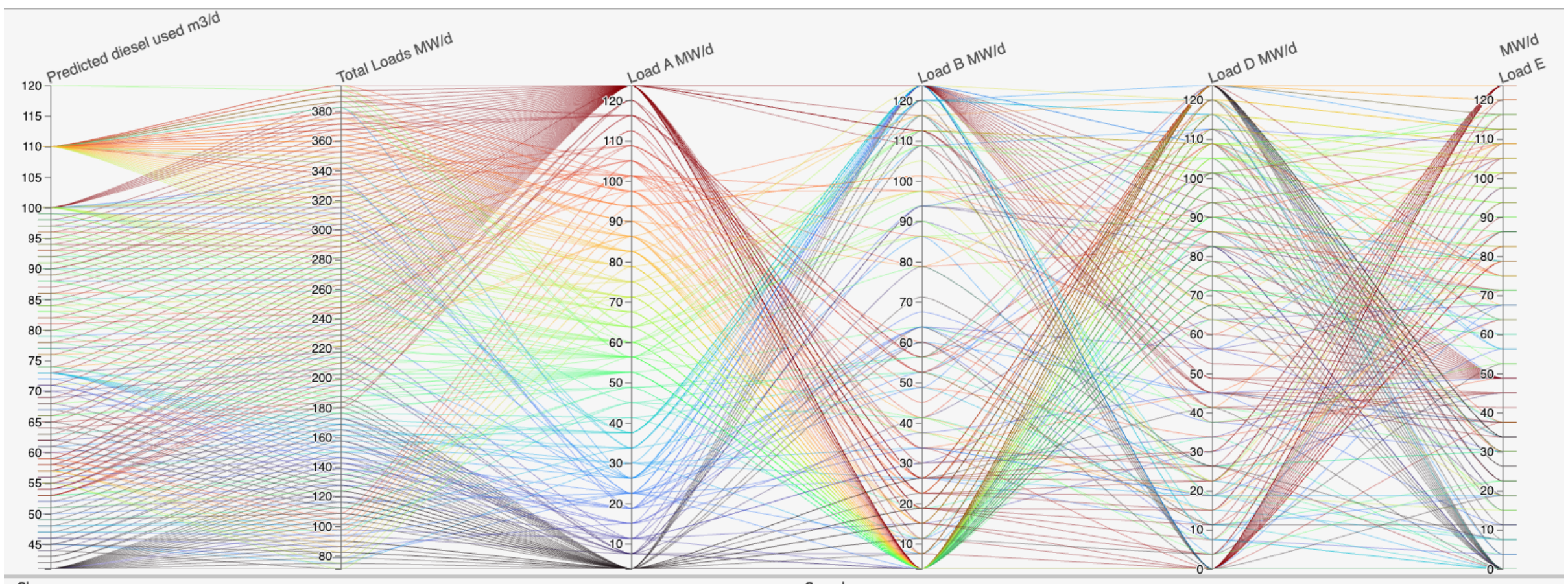

Fig. 8. Parallel Plot Coordinate with multiple axes representing the daily power loads of the Diesel Generators (DGs) individually and as a sum along with the predicted diesel consumed per day.

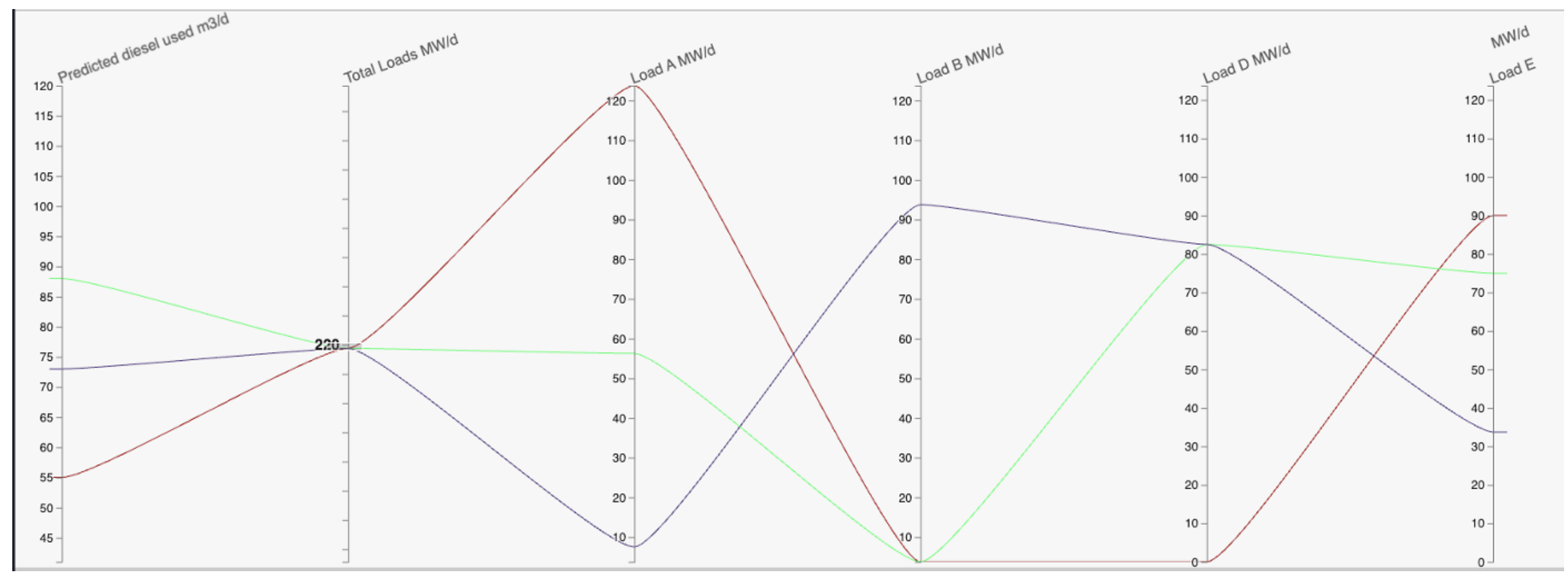

| Color | Predicted diesel used | Total Loads MW/d | Load A MW/d | Load B MW/d | Load D MW/d | Load E MW/d |
|---|---|---|---|---|---|---|
| | 88 | 217.5 | 56.25 | 3.75 | 82.5 | 75 |
| | 73 | 217.5 | 7.5 | 93.75 | 82.5 | 33.75 |
| | 55 | 217.5 | 123.75 | 3.75 | 0 | 90 |

Fig. 9. The best, median and worst daily load combinations with their respective diesel use per day according to the total loads needed that day.

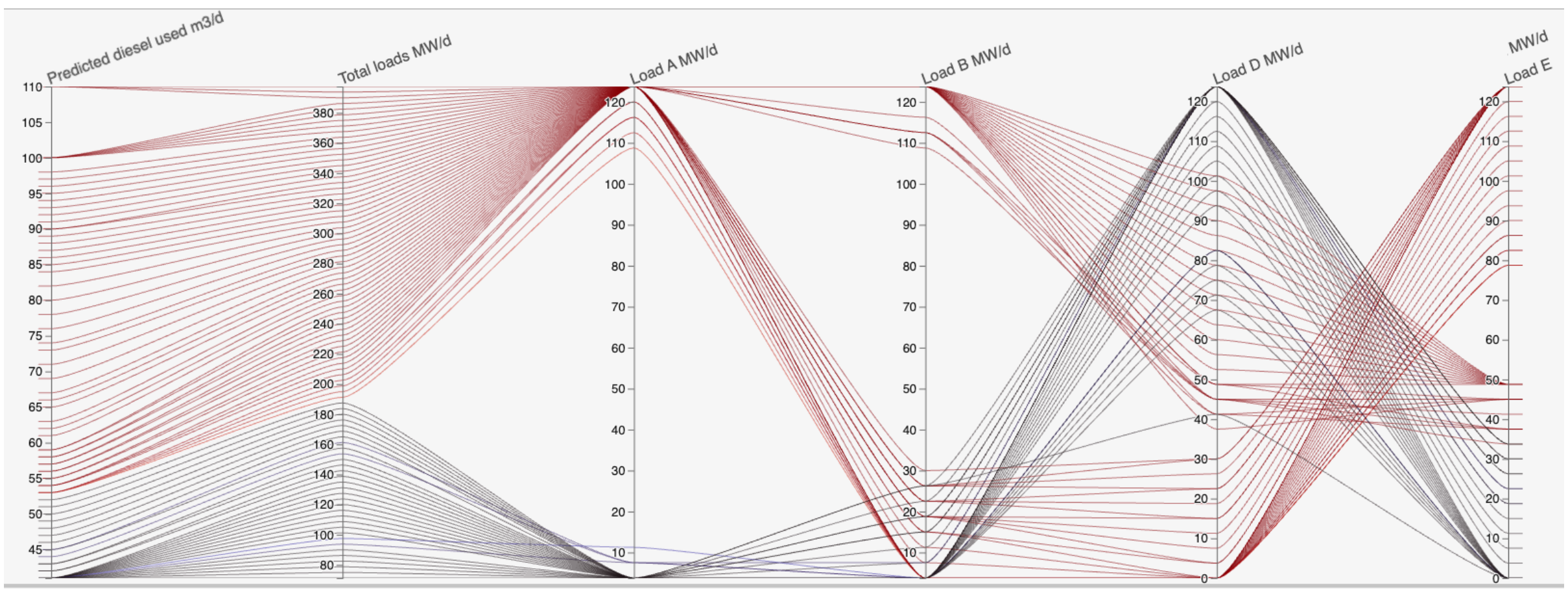

Fig. 10. Parallel Plot Coordinate for the combinations that used the least amount of diesel for the required daily total loads with the black lines representing low levels of diesel used and low total loads levels per day and the red lines representing relatively high diesel used and high total loads per day.

In Figure 10 and Supplementary Figures 1 to 9, a clear linear correlation emerges between daily total diesel consumption and the corresponding required daily loads. This correlation aligns with the intuitive expectation that higher power demands necessitate greater fuel usage. Notably, when transitioning from approximately 70 MW to 130 MW in daily total loads, the most efficient DG combination maintained consistent daily diesel consumption. This consistency indicated that within this range, optimisation of daily diesel usage becomes challenging due to minimal fuel consumption and operational constraints on the DGs. Furthermore, as daily total loads increased, the highest or lowest loads on individual DGs remained unchanged for each combination with a daily total load increase of less than 20 MW. However, with variations exceeding 20 MW in daily total loads, the highest and lowest power loads on individual DGs fluctuated.

The variations in diesel consumption according to the different combinations of loads occured because each of the DGs has different rated power and efficiency. The efficiency of a DG is inversely proportional to the rated power, fuel consumption rate and $CO_2$ emissions of the DG (Jakhrani et al., 2012). Therefore, the rated power of the selected DG should be close to the load demand. If the power load of the DG is close to the rated power of the DG, the DG will be most efficient. However, if the power load is too high compared to the rated power, the DG will be overloaded which results in poor efficiency, and a high specific fuel consumption rate and can even damage the DG. If the power loads are low compared to the rated power of the DGs they will operate at lower efficiency which can lead to high fuel consumption and increased maintenance costs. However, it is to be noted that the effect of the power factor also plays a role in the efficiency and rated power of the DG. When selecting a DG, it is important to consider the expected load profile and choose a DG with a power factor that is appropriate for the type of loads that will be connected to it.

Although determining the rated power output for each DG and developing a simple mathematical model to find the optimal loading is possible, the complexity of the system and the interaction of multiple pieces of equipment on the oil platform results in the need to have a thorough technical understanding of the system itself to proceed. Instead, the ML model is able to capture various arising, complex and non-linear relationships which may be affecting diesel consumption, other than the loads on the DGs, such as IGGs, PFHs, changing weather conditions, and equipment ageing during longer periods. Additionally, other factors which may have an influence on diesel consumption which is taken into account in the ML model include DG types and operative settings, maintenance schedules and operational restraints. The ML model can easily adapt to changing patterns and search for possible relationships if it is trained on a sufficiently large and representative set of data, a condition that may not apply to a simple static mathematical model. Moreover, the ML model can perform more accurate predictions for better planning and resource allocation and can be scaled to accommodate future needs.

### 3.3. Fuel Savings

Figure 11 displays the diesel consumed for the different combinations of the four DGs which consumed the least, medium and most amount of diesel for a range of total required daily power loads.

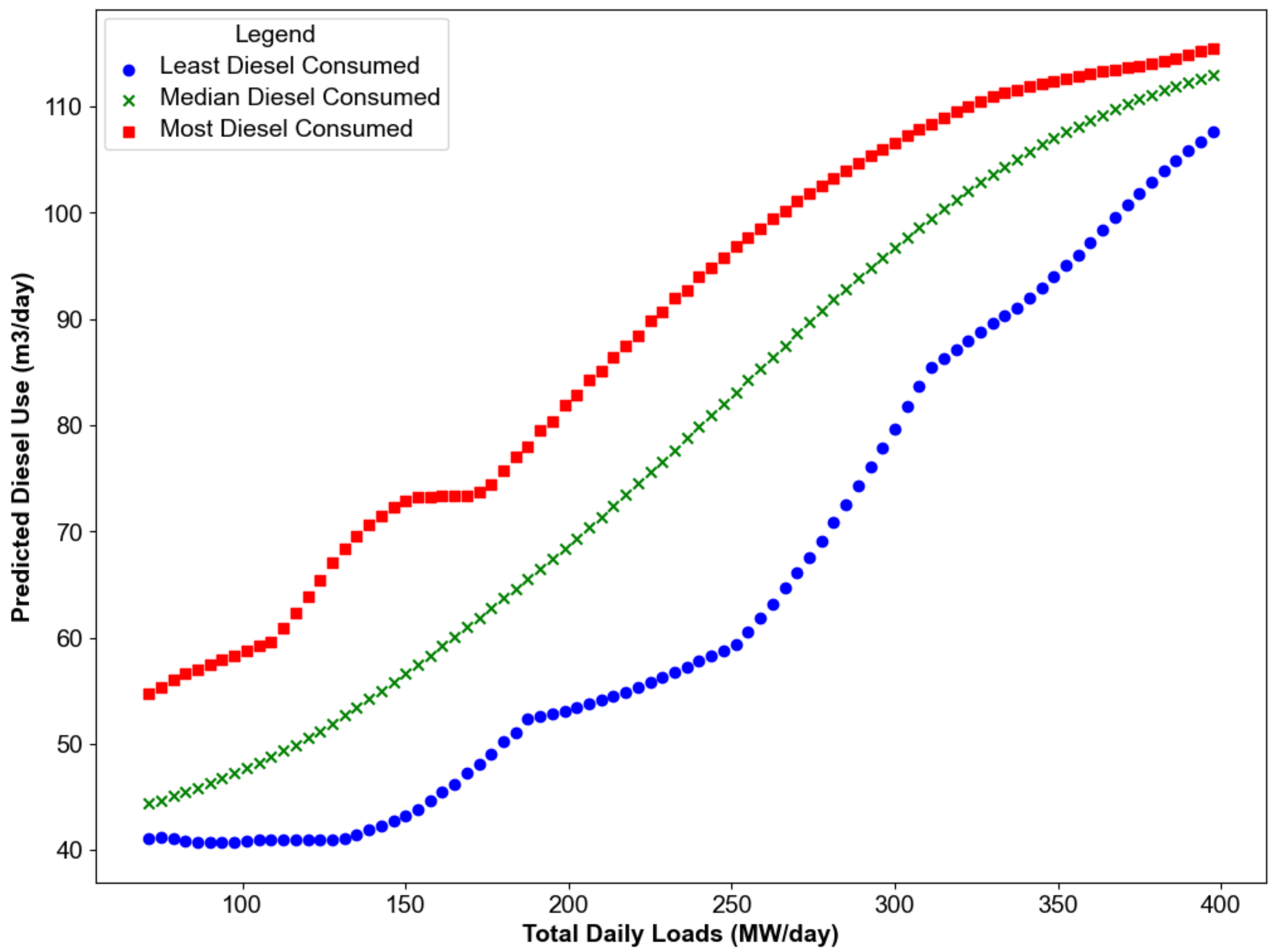

Fig. 11 Predicted diesel used against daily total loads for the Diesel Generators (DGs) combinations which used the least, medium and most amount of diesel.

From Figure 11, if an average of the daily diesel savings was calculated when using the best-case scenario where the combinations of power loads which used the least amount of diesel were considered instead of using the worst-case scenario where the combination of power loads which used the most amount of diesel was considered, it results in approximately 24 $m^3$/day. On the other hand, the average diesel savings that can be obtained when using the best-case power loads combinations scenario (where the least amount of diesel is consumed) instead of the median power loads combinations scenario (where a median amount of diesel is consumed) is approximately 17 $m^3$/day. For example, the fuel savings from literature such as Knudsen et al., (2017) a Genetic Algorithm outperformed an even distribution approach by 161 litres/hour (3.9 $m^3$/day) and 394 litres/hour (9.5 $m^3$/day) for the most unfavourable choice. Even though, this past study used datasets and methods different from the present study, comparing the savings indicates that this study allowed relatively higher fuel savings to be made in the same context. Moreover, considering the original real-world dataset used in this study for training and testing the model, the average daily diesel consumption was approximately 74 m3/day over 18 months. In the best-case scenario, where load combinations used the least amount of diesel, the resulting average consumption was about 65 $m^3$/day. This leads to potential savings of 9 $m^3$/day or 9000 litres/day, representing an overall approximate savings of 12% compared to the average diesel consumption in the real-world dataset.

However, it is important to note that the average predicted daily diesel consumption for the best case scenario (from the power loads combinations consuming the least amount of diesel) was calculated using generated data, with more rows (data points) than the actual recorded data from the oil platform. Acknowledging that the generated data is bigger than the actual real-world dataset leads to the belief that more actual recorded data would enhance model accuracy and reliability of the savings being calculated. The daily fuel savings were depicted in Figure 12, indicating a peak of approximately 250 MW of daily total power. This suggests that an optimal daily fuel savings point can be achieved at this specific level of daily total power loads.

The variations in daily fuel (diesel) savings were graphically represented in Figure 12 as the daily total loads increased. Notably, the fuel savings reached their peak at approximately 250 MW of total power daily before exhibiting a decline. This observation underscores the idea that an optimal level of daily fuel savings can be achieved specifically at this particular amount of daily average total loads.

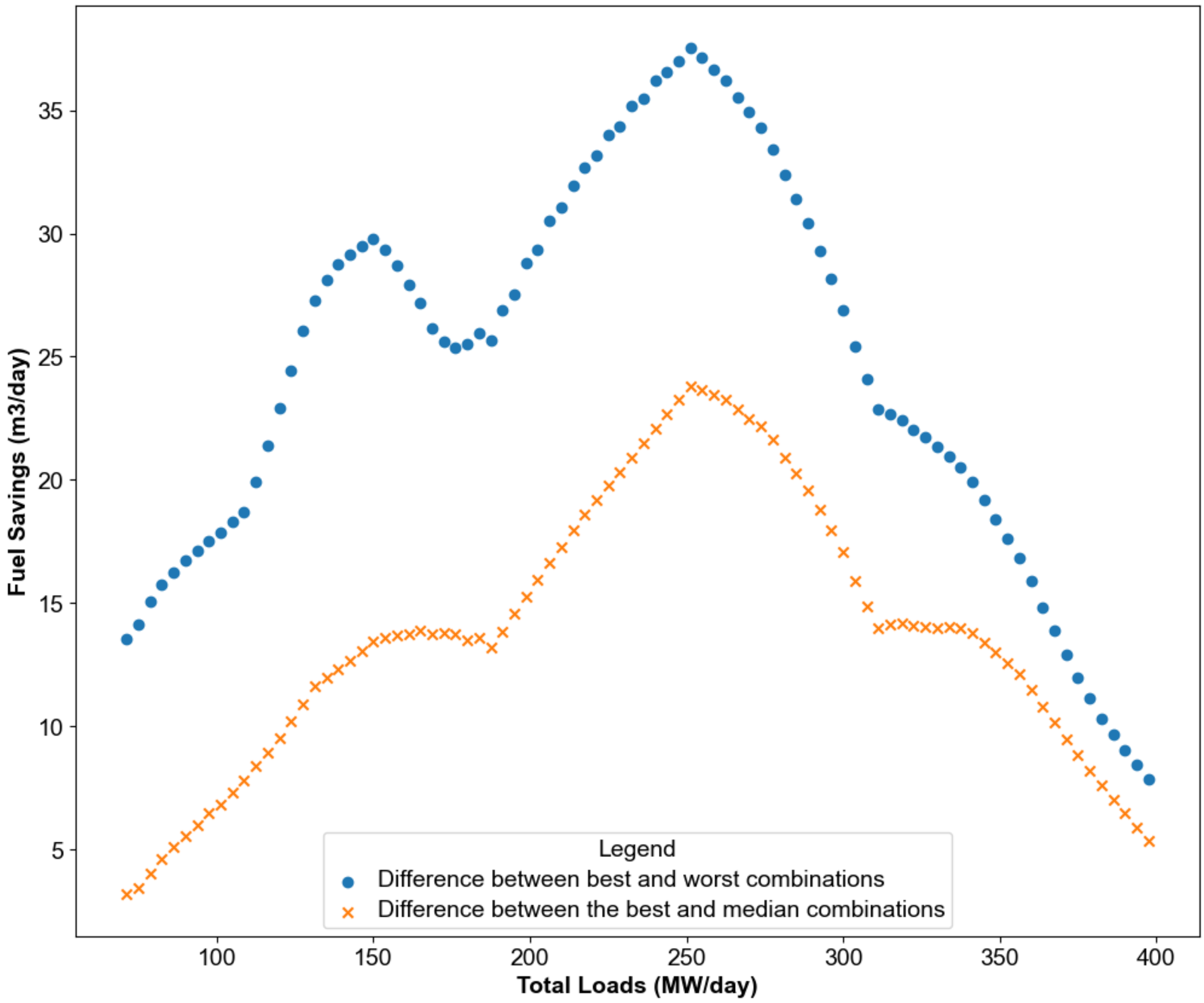

Fig. 12. Fuel savings against daily total power loads.

## 4. Conclusion

This research explored a lesser-known approach for mitigating diesel consumption on offshore oil platforms by introducing a tailored ML methodology for optimising the loading of DGs. This study utilised data sourced from an FPSO oil platform in Scotland, encompassing varying time intervals. The initial correlation of 0.68 between daily average power loads for multiple DGs and diesel consumption improved to 0.86 after meticulous preprocessing and outlier removal, further highlighting the need for preprocessing. Among various ML models explored, MLR and ANN emerged as the most effective in predicting diesel consumption among others such as RF, XGBoost and ETRs. Using the best performing ML model, input features were generated and diverse power loads combinations were explored for the DGs. The best load combination, which minimised predicted daily diesel consumption, was identified through a brute force exhaustive exploration. To contribute to this field of work, the findings demonstrated the possibility of significant fuel savings with an approximate 21 % reduction achievable when calculating an average of the best load combinations (consuming the least amount of daily diesel) instead of using an average of median load combinations. Furthermore, opting for the best load combinations (one which consumed the least amount of daily diesel) over the worst (one which consumed the most amount of daily diesel) yielded an approximate 27 % reduction in fuel consumption. The identification of optimal loading strategies for DGs has therefore been well established using the present study. Moreover, the development of an interactive visualisation tool, that can be used by industries, to determine the corresponding combinations of power load configurations for the DGs using the least, medium and the most amount of daily diesel from daily total loads values that are anticipated to be required on a particular day, further illustrated the importance of emphasising on the applicability of such studies being conducted in research. Although, this research ended in a practical outcome, the limitations of this work, can include the reliance on a limited and not having a fully representative dataset. Further exploration with substantial data sets is necessary to enhance the robustness and generalisability of this methodology. The applicability of the findings to diverse oil platforms with different DGs also requires further investigation.

## 5. Data Availability

Datasets related to this article belong to a specific oil platform in Scotland .

## 6. Declaration of Competing Interest

The authors declare that they have no known competing financial interests or personal relationships that could have appeared to influence the work reported in this paper.

## 7. Acknowledgements

This work was supported by UK Research and Innovation (UKRI) through EPSRC research grant number EP/S022996/1. The authors extend their appreciation to Intelligent Plant for also funding this work. I. Triguero holds a Maria Zambrano Fellowship at the University of Granada, and his work is also partly supported by the Spanish projects A-TIC-434-UGR20 and PID2020-119478GB-I00.

## 8. References

Ahlgren, F., & Thern, M. (2018). *Auto Machine Learning for predicting Ship Fuel Consumption*. https://www.python.org/

Allahyarzadeh, A., Jonas Dezan, D., Oliveira Salviano, L., de Oliveira Junior, S., Allahyarzadeh-Bidgoli, A., de Oliveira Junior, S., & Itizo Yanagihara, J. (2018). *Optimization Procedure to Minimize FPSO Fuel Consumption under Two Operation Modes in a Brazilian Deep-Water Oil Field*. https://www.researchgate.net/publication/325877708

Barbosa, L. F. F. M., Nascimento, A., Mathias, M. H., & de Carvalho, J. A. (2019). Machine learning methods applied to drilling rate of penetration prediction and optimization - A review. *Journal of Petroleum Science and Engineering*, *183*. https://doi.org/10.1016/j.petrol.2019.106332

Bijan, K., Juan Carlos, G., Gustavo, L. T., Teresa, E. P., & Claudio, M. (2005). *Materials Optimisation In Hydrocarbon Production* . https://onepetro.org/NACECORR/proceedings-abstract/CORR05/All-CORR05/115132

Cemeljic, H., Havelka, J., Jeremic, A., & Kuzle, I. (2024). Adaptive algorithm for controlling the power management system on offshore jack-up drilling rigs. *Energy Sources, Part A: Recovery, Utilization, and Environmental Effects, 46*(1), 16717-16737. https://doi.org/10.1080/15567036.2024.2430415

Chaudhary, A. (2019). *Developing predictive models for fuel consumption and maintenance cost using equipment fleet data.*

Choubey, S., & Karmakar, G. P. (2021). Artificial intelligence techniques and their application in oil and gas industry. *Artificial Intelligence Review*, *54*(5), 3665–3683. https://doi.org/10.1007/s10462-020-09935-1

CK Power. (2023). *Keys to running your generator efficiently _ CK Power*.

EIA. (2023). *Short-Term Energy Outlook Global oil markets*. https://www.eia.gov/outlooks/steo/report/global_oil.php

Facebook AI Research. (2020). *HiPlot demonstration — HiPlot 0.1.33 documentation*. Github. https://facebookresearch.github.io/hiplot/

Gong, J., Shang, J., Li, L., Zhang, C., He, J., & Ma, J. (2021). A comparative study on fuel consumption prediction methods of heavy-duty diesel trucks considering 21 influencing factors. *Energies*, *14*(23). https://doi.org/10.3390/en14238106

Halpin, T., & Morgan, T. (2008). Relational Languages. In *Information Modeling and Relational Databases* (pp. 527–635). Elsevier. https://doi.org/10.1016/b978-012373568-3.50016-3

Hannah, R., & Max, R. (n.d.). *Electricity Mix - Our World in Data*. Our World in Data. Retrieved February 20, 2023, from https://ourworldindata.org/electricity-mix

Hariz, H. ben, & Thesis, P. (2010). *The Optimisation of the Usage of Gas Turbine Generation Sets for Oil and Gas Production Using Genetic Algorithms*.

Hegde, C., & Gray, K. E. (2017). Use of machine learning and data analytics to increase drilling efficiency for nearby wells. *Journal of Natural Gas Science and Engineering*, *40*, 327–335. https://doi.org/10.1016/j.jngse.2017.02.019

ITHACA energy. (n.d.). *Captain Ithaca Energy*.

J. D. Jobson. (1991). Multiple Linear Regression. *Applied Multivariate Data Analysis. Springer Texts in Statistics. Springer*. Jobson, J.D. (1991). Multiple Linear Regression. In: Applied Multivariate Data Analysis. Springer Texts in Statistics. Springer, New York, NY. https://doi.org/10.1007/978-1-4612-0955-3_4

JavaTpoint. (2021). *Brute force approach*. Www.Javatpoint.Com. https://www.javatpoint.com/brute-force-approach

Keras. (2023). *Keras*. https://keras.io/

Khan, L., Mumtaz, S., & Khattak, A. (2012). Comparison of Genetic Algorithm and Harmony Search for Generator Maintenance Scheduling. *Mehran University Research Journal of Engineering & Technology, 31*(4). https://doi.org/10.06.2010

Khatib, T., Mohamed, A., Sopian, K., & Mahmoud, M. (2011). Optimal sizing of building integrated hybrid PV/diesel generator system for zero load rejection for Malaysia. *Energy and Buildings*, *43*(12), 3430–3435. https://doi.org/10.1016/j.enbuild.2011.09.008

Khor, C. S., Elkamel, A., & Shah, N. (2017). Optimization methods for petroleum fields development and production systems: a review. *Optimization and Engineering*, *18*(4), 907–941. https://doi.org/10.1007/s11081-017-9365-2

Knudsen, J., Bendtsen, J., Andersen, P., Madsen, K., Sterregaard, C., & Rossiter, A. (2017). Fuel optimization in multiple diesel driven generator power plants. *1st Annual IEEE Conference on Control Technology and Applications, CCTA 2017*, *2017-January*, 493–498. https://doi.org/10.1109/CCTA.2017.8062510

Kusakana, K. (2015). Optimisation of the daily operation of a hydrokinetic-diesel generator power plant. *Proceedings of the Conference on the Industrial and Commercial Use of Energy, ICUE*, *2015-September*, 367–372. https://doi.org/10.1109/ICUE.2015.7280291

Lou, Q. H., Niu, H. H., Chen, J., Gao, Y., Geng, X., & Li, B. (2019). Research on the analysis and optimization method of offshore oil-platforms energy system. *IOP Conference Series: Earth and Environmental Science*, *267*(6). https://doi.org/10.1088/1755-1315/267/6/062052

*Mahalanobis Distance*. (2008).

Meng, Q., Guo, M., Zhao, R., & Wang, W. (2025). Research on low carbon power system planning method for offshore oilfield considering wind power and carbon capture, utilization and storage. *Electric Power Systems Research, 248*, 111907. https://doi.org/10.1016/j.epsr.2025.111907

Mobarra, M., Issa, M., Rezkallah, M., & Ilinca, A. (2019). Performance Optimization of Diesel Generators Using Permanent Magnet Synchronous Generator with Rotating Stator. *Energy and Power Engineering*, *11*(07), 259–282. https://doi.org/10.4236/epe.2019.117017

Nguyen, T. van, Barbosa, Y. M., da Silva, J. A. M., & de Oliveira Junior, S. (2019). A novel methodology for the design and optimisation of oil and gas offshore platforms. *Energy*, *185*, 158–175. https://doi.org/10.1016/j.energy.2019.06.164

Nguyen, T. van, Voldsund, M., Breuhaus, P., & Elmegaard, B. (2016). Energy efficiency measures for offshore oil and gas platforms. *Energy*, *117*, 325–340. https://doi.org/10.1016/j.energy.2016.03.061

Peixoto, C. S., Vieira, G. G. T. T., Salles, M. B. C., & Carmo, B. S. (2024). Assessing the impact of power dispatch optimization and energy storage systems in diesel-electric PSVs: A case study based on real field data. *Applied Energy, 357*, 122476. https://doi.org/10.1016/j.apenergy.2023.122476

Rahmanifard, H., & Plaksina, T. (2019). Application of artificial intelligence techniques in the petroleum industry: a review. In *Artificial Intelligence Review* (Vol. 52, Issue 4, pp. 2295–2318). Springer Netherlands. https://doi.org/10.1007/s10462-018-9612-8

Rao, K. S., Chauhan, P. J., Panda, S. K., Wilson, G., Liu, X., & Gupta, A. K. (2015). Optimal scheduling of diesel generators in offshore support vessels to minimize fuel consumption. *IECON 2015 - 41st Annual Conference of the IEEE Industrial Electronics Society*, 4726–4731. https://doi.org/10.1109/IECON.2015.7392838

Riayatsyah, T. M. I., Geumpana, T. A., Fattah, I. M. R., & Mahlia, T. M. I. (2022). Techno-Economic Analysis of Hybrid Diesel Generators and Renewable Energy for a Remote Island in the Indian Ocean Using HOMER Pro. *Sustainability (Switzerland)*, *14*(16). https://doi.org/10.3390/su14169846

Roberts, R. C., Laramee, R. S., Smith, G. A., & Brookes, P. (2015). *Smart Brushing for Parallel Coordinates* (Vol. 14, Issue 8).

*Role of Diesel Power Generators in the Oil & Gas Industry*. (2023). https://www.generatorsource.com/

SAS. (n.d.). *Data Mining Software, Model Development and Deployment, SAS Enterprise Miner _ SAS UK*. Retrieved March 22, 2023, from https://www.sas.com/en_gb/software/enterprise-miner.html

Silva, T. L., & Camponogara, E. (2014). A computational analysis of multidimensional piecewise-linear models with applications to oil production optimization. *European Journal of Operational Research*, *232*(3), 630–642. https://doi.org/10.1016/j.ejor.2013.07.040

Sircar, A., Yadav, K., Rayavarapu, K., Bist, N., & Oza, H. (2021). Application of machine learning and artificial intelligence in oil and gas industry. In *Petroleum Research* (Vol. 6, Issue 4, pp. 379–391). KeAi Publishing Communications Ltd. https://doi.org/10.1016/j.ptlrs.2021.05.009

Sklearn. (2023a). *scikit-learn 1.2.2 documentation*. https://scikit-learn.org/stable/modules/generated/sklearn.linear_model.LinearRegression.html

Sklearn. (2023b). *sklearn.ensemble.ExtraTreesRegressor*.

*Spearman Rank Correlation Coefficient*. (2008).

Svendsen, H. G. (2022). *Optimised operation of low-emission offshore oil and gas platform integrated energy systems*. http://arxiv.org/abs/2202.05072

TensorFlow. (2023). *TensorFlow*. https://www.tensorflow.org/

Wickramanayake, S., & Bandara, D. H. M. N. (2016). Fuel consumption prediction of fleet vehicles using Machine Learning: A comparative study. *2nd International Moratuwa Engineering Research Conference, MERCon 2016*, 90–95. https://doi.org/10.1109/MERCon.2016.7480121

World Nuclear Association. (n.d.). *Where does our electricity come from.*

Worldwide Power Products. (2023). *Power Management Options For Offshore Oil & Gas Rigs*.

XGBoost Developers. (2022). *XGBoost Documentation*. https://xgboost.readthedocs.io/en/stable/

Yadav, P., Kumar, R., Panda, S. K., & Chang, C. S. (2011). An Improved Harmony Search algorithm for optimal scheduling of the diesel generators in oil rig platforms. *Energy Conversion and Management*, *52*(2), 893–902. https://doi.org/10.1016/j.enconman.2010.08.016

Zhang, A., Zhang, H., Qadrdan, M., Yang, W., Jin, X., & Wu, J. (2019). Optimal planning of integrated energy systems for offshore oil extraction and processing platforms. *Energies*, *12*(4). https://doi.org/10.3390/en12040756

Zheng, H., Zhou, H., Kang, C., Liu, Z., Dou, Z., Liu, J., Li, B., & Chen, Y. (2021). Modeling and prediction for diesel performance based on deep neural network combined with virtual sample. *Scientific Reports*, *11*(1). https://doi.org/10.1038/s41598-021-96259-x

Zheng, X., SHI, J., CAO, G., YANG, N., CUI, M., JIA, D., & LIU, H. (2022). Progress and prospects of oil and gas production engineering technology in China. *Petroleum Exploration and Development*, *49*(3), 644–659. https://doi.org/10.1016/S1876-3804(22)60054-5